\documentclass[runningheads]{llncs}
\usepackage[T1]{fontenc}
\usepackage{graphicx}
\usepackage{tabularray}
\SetTblrInner{rowsep=0pt}
\usepackage{amsmath} 
\usepackage{amssymb} 
\usepackage{hyperref}
\usepackage{color}
\usepackage[misc]{ifsym}

\begin{document}
%
\title{KARGEN: Knowledge-enhanced Automated Radiology Report Generation Using Large Language Models}
\titlerunning{KARGEN: Knowledge-enhanced R2Gen Using LLMs}
%
\author{Yingshu Li\inst{1} \and
Zhanyu Wang\inst{1} \and
Yunyi Liu\inst{1}  \and
Lei Wang\inst{2}  \and
Lingqiao Liu\inst{3}  \and
Luping Zhou\inst{1(\textrm{\Letter})}}  
%
\authorrunning{Y. Li et al.}
%
\institute{University of Sydney, Sydney, NSW, Australia \\ \email{\{yingshu.li, zhanyu.wang, yunyi.liu1, luping.zhou\}}@sydney.edu.au \and
University of Wollongong, Wollongong, NSW, Australia \\
\email{leiw@uow.edu.au}\\  \and
The University of Adelaide, Australia\\
\email{lingqiao.liu@adelaide.edu.au}}
\maketitle              
\renewcommand{\thefootnote}{}
\footnotetext{\inst{\textrm{\Letter}}Corresponding Author}
\renewcommand{\thefootnote}{\arabic{footnote}}
\setcounter{footnote}{0}
\begin{abstract}Harnessing the robust capabilities of Large Language Models (LLMs) for narrative generation, logical reasoning, and common-sense knowledge integration, this study delves into utilizing LLMs to enhance automated radiology report generation (R2Gen). Despite the wealth of knowledge within LLMs, efficiently triggering relevant knowledge within these large models for specific tasks like R2Gen poses a critical research challenge. This paper presents KARGEN, a \textbf{K}nowledge-enhanced \textbf{A}utomated radiology \textbf{R}eport \textbf{GEN}eration framework based on LLMs. Utilizing a frozen LLM to generate reports, the framework integrates a knowledge graph to unlock chest disease-related knowledge within the LLM to enhance the clinical utility of generated reports. This is achieved by leveraging the knowledge graph to distill disease-related features in a designed way. Since a radiology report encompasses both normal and disease-related findings, the extracted graph-enhanced disease-related features are integrated with regional image features, attending to both aspects. We explore two fusion methods to automatically prioritize and select the most relevant features. The fused features are employed by LLM to generate reports that are more sensitive to diseases and of improved quality. Our approach demonstrates promising results on the MIMIC-CXR and IU-Xray datasets. Our code will be available on GitHub.

\keywords{Radiology Report Generation  \and Medical Domain Knowledge Graph \and Large Language Models.}
\end{abstract}
\section{Introduction}

Automated radiology report generation (R2Gen) is gaining traction due to its potential to streamline the time-consuming and error-prone task of medical image reading and report writing. Unlike generic image captioning tasks~\cite{xu2015show,lu2017knowing,cornia2020meshed}, which focus on concise summaries of image contents, R2Gen involves generating detailed paragraphs covering both normal and pathological findings in radiology images. Various approaches address this challenge~\cite{wang2021self,chen2020generating,chen2022cross}. For instance, hierarchically structured LSTM~\cite{yin2019automatic,wang2021self} and memory-driven modules~\cite{chen2020generating,chen2022cross} enhance long-term memory capabilities. Data deviation, where normal contents dominate, is another challenge~\cite{wang2023rethinking,liu2021exploring}. Efforts to tackle this involve improving image-text attention, aligning features, and incorporating external domain knowledge~\cite{wang2021self,wang2022medical,liu2021exploring}. Some studies~\cite{wang2022medical} leverage additional disease classification tasks, while others~\cite{liu2021exploring,yang2022knowledge} utilize knowledge graphs to capture disease-related information based on medical domain knowledge.

In the past two years, large language models (LLMs)~\cite{touvron2023llama} have demonstrated significant capabilities in generating more human-like, coherent, and contextually relevant responses, utilizing their extensive knowledge base. This potential has also been explored to combat the aforementioned challenges for R2Gen~\cite{wang2023r2gengpt,pellegrini2023radialog}. However, despite the wealth of knowledge within LLMs, efficiently triggering relevant knowledge within these large models for specific tasks like R2Gen could pose a critical research challenge. Current methods, relying primarily on visual prompts from regional image features, may struggle to capture detailed, disease-related information to effectively prompt LLMs for R2Gen. Although~\cite{pellegrini2023radialog} trained a disease classifier and constructed its output as an additional text prompt, the information provided remains arguably sparse as clues for diseases.

In this paper, we present KARGEN, a novel Knowledge-enhanced Automated radiology Report GENeration framework based on LLMs. To the best of our knowledge, this is the first exploration of integrating a disease-specific knowledge graph to activate and unlock pertinent medical domain knowledge within LLMs. Diverging from previous approaches that constructed graph convolutional networks (GCNs) solely based on image or text features, our method integrates both text and image features to define graph nodes, linking regional image features with the text embedding of disease classes. Our approach fosters a comprehensive fusion of inter-disease features, allowing us to capture fine-grained disease-related features and interrelationships among diseases. Moreover, unlike prior methods that merely use graph-enhanced features for R2Gen, we advocate for the integration of both graph-enhanced disease-related features and regional image features to attend to both normal and disease-related findings in a radiology report. We therefore develop two fusion methods, operating at either individual feature element or modality (feature types) level, to effectively prioritize the most relevant features. These fused features are then leveraged to prompt LLMs to generate reports to become more sensitive to diseases and achieve improved quality.

Our main contributions are summarized as follows:
\begin{itemize}
    \item[(1)] We present a novel framework that integrates a medical domain knowledge graph with LLMs for R2Gen. It demonstrates, for the first time, that despite the wealth of knowledge within LLMs, the incorporation of a specific knowledge graph encoding disease information is necessary and beneficial for activating relevant knowledge in LLMs for R2Gen.
    \item[(2)] Our model includes a novel knowledge graph for extracting disease-related features, along with two alternative strategies for a feature fusion component. These strategies effectively integrate graph-enhanced disease-related features and regional image features, enabling the model to attend to both normal and disease-related content in the generated reports.
    \item[(3)] Our method, validated on two public datasets IU-Xray and MIMIC-CXR, outperforms multiple relevant state-of-the-art methods on various evaluation metrics including the most recent clinic-related ones.
\end{itemize}

\section{Methodology}
\begin{figure}[t]
\centering
\includegraphics[width=\textwidth]{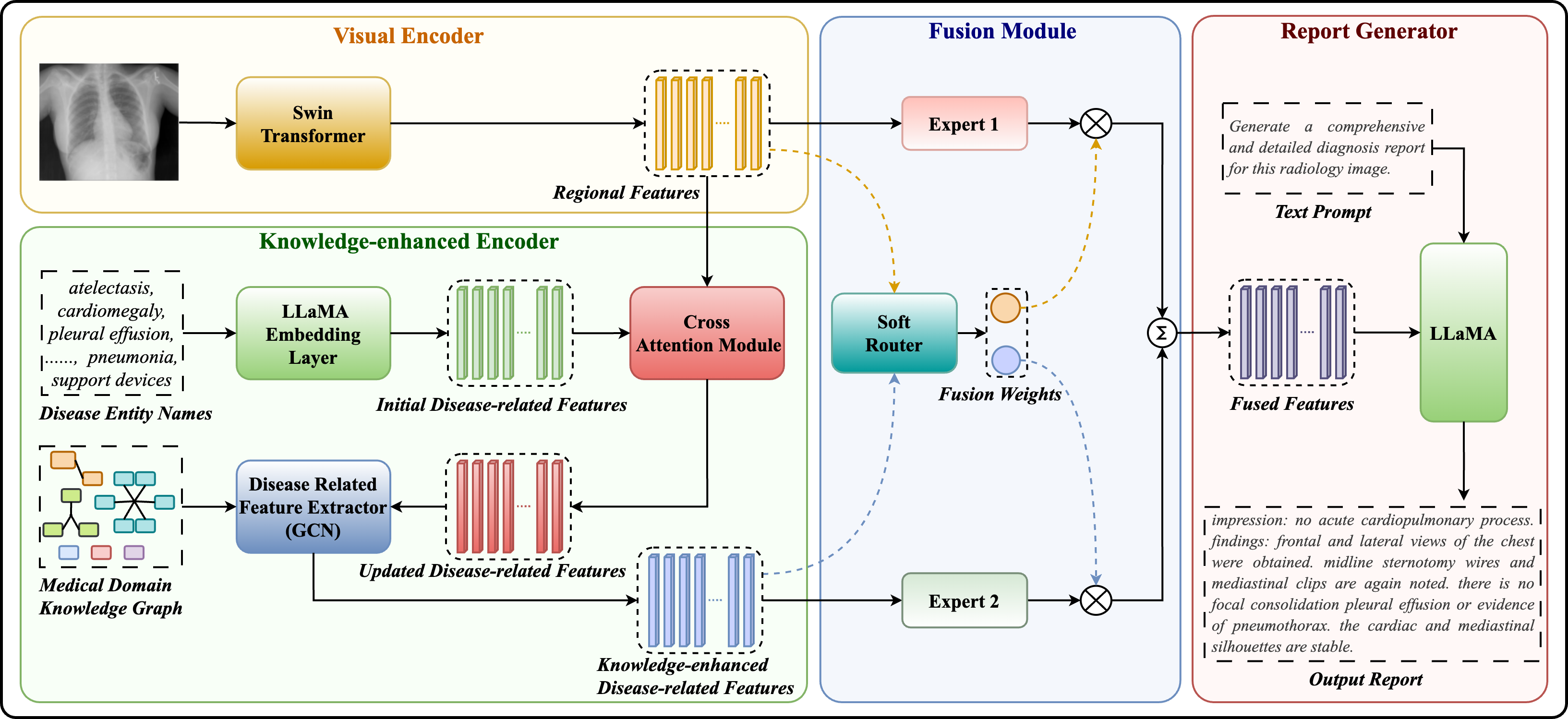}
\caption{An overview of the KARGEN framework, which comprises a visual encoder, a knowledge-enhanced encoder, a fusion module and a report generator.} 
\label{Figs: overall}
\end{figure}

Our framework consists of four main components: a visual feature encoder, a knowledge-enhanced feature encoder, a feature fusion module, and a report generator. The visual feature encoder extracts regional features from a chest x-ray image, which are then processed by the knowledge-enhanced feature encoder to `distill' disease-related information guided by a medical knowledge graph. The resulting knowledge-enhanced disease-related features are fused with the regional image features in the feature fusion module and used to prompt the LLaMA-based report generator for R2Gen. Fig.~\ref{Figs: overall} gives an overview of KARGEN. 
\subsection{Feature Extraction}
\noindent \textbf{Regional Feature Extraction}~~~Given an input X-ray image $\mathbf{X}_v$, we initially extract regional image features $\mathbf{Z}_v = Swin(\mathbf{X}_v; \theta_v)$, utilizing a pre-trained Swin Transformer~\cite{liu2021swin}, where $\mathbf{Z}_v \in \mathbb{R}^{S \times d_v}$ ($S$: the number of features; $d_v$: the dimensionality of each feature; $\theta_v$: the parameters of the Swin Transformer). 

\begin{figure}[h]
\centering
\includegraphics[width=0.7\textwidth]{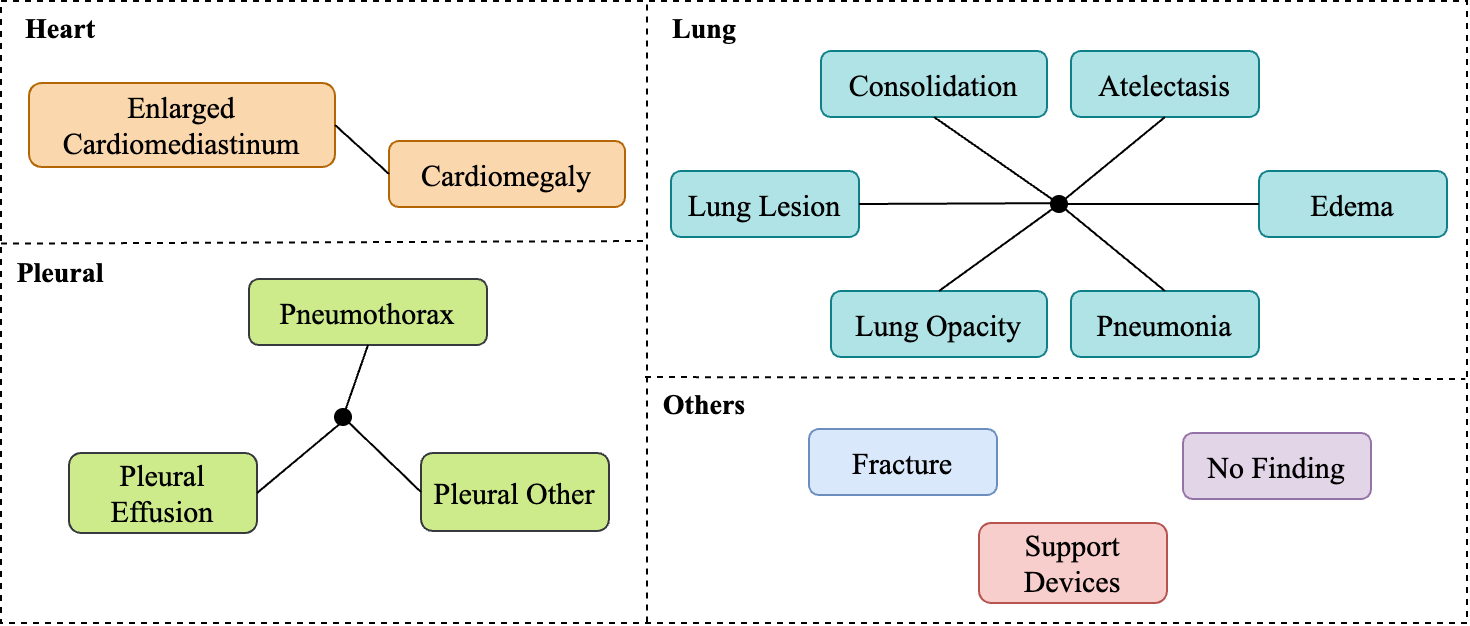}
\caption{The medical domain knowledge graph is constructed based on the correlations among various diseases, where diseases that are linked together are interconnected.} 
\label{Figs: KG}
\end{figure}

\noindent \textbf{Medical Domain Knowledge Graph}~~~Focusing on disease-related features in medical imaging, especially for interrelated chest diseases, is critical. We propose a medical domain knowledge graph to extract chest disease features, incorporating 14 terms from the Chexpert~\cite{irvin2019chexpert}. Each disease entity is represented by the word embedding of its name, obtained using the LLaMA Word Embedding Layer. The connections are illustrated in Fig.~\ref{Figs: KG},  highlighting that abnormalities within the same region exhibit stronger correlations than those across different organs. This guides our analysis of diseases in the lungs, heart, and pleura in chest X-ray images~\cite{zhang2020radiology}, capturing nuanced relationships effectively. 

\noindent \textbf{Diseased-related Feature Extraction}~~~
Based on the knowledge graph, we construct a GCN to aggregate disease-related features.  Our GCN comprises three layers. At each layer $l$ ($l=1,2 \ \text{and} \ 3$), there are two primary phases: 1) the propagation of information throughout the graph, and 2) the updating of disease-related features. Let $\mathbf{N}^l$ denote the node features in the $l$-th layer, $\mathbf{A}$ denote the adjacency matrix governed by the knowledge graph, and $\mathbf{A}' = \mathbf{D}^{-1/2} \cdot \mathbf{A} \cdot \mathbf{D}^{-1/2}$ ($\mathbf{D}$ is the degree matrix of $\mathbf{A}$). The entire process can be formulated as
\begin{align}
&\mathbf{N}^{l}_{phase 1} = GELU(LN((\mathbf{N}^{l-1} \cdot \mathbf{W}^l)\mathbf{A}')), \\\nonumber
&\mathbf{N}^{l} = GELU(LN((\mathbf{N}^{l-1} + \mathbf{N}^l_{phase 1}) \cdot \mathbf{W}_{update}^l + \mathbf{N}^{l-1})).
\end{align}
Here $\mathbf{W}^l$ and $\mathbf{W}_{update}^l$ are learnable parameters for information propagation and updating. $LN(\cdot)$ denotes variants of layer normalization.

The initial node features $\mathbf{N}^{0} $ are determined by using the disease entity name embeddings $\mathbf{E} \in \mathbb{R}^{M \times d_{w}} = [\mathbf{e}_1,\cdots,\mathbf{e}_i,\cdots,\mathbf{e}_M]^T$($M=14$) to query the regional image features $\mathbf{Z}_v$ output by the visual encoder via multi-head attention:
\begin{equation}
\begin{gathered}
\mathbf{N}^0 = MHA(\mathbf{E}, \mathbf{Z}_v) = Concat(head_1, ..., head_h)\mathbf{W}^O, \\
head_h = Softmax\left(\frac{\mathbf{Q}_h\mathbf{K}_h^T}{\sqrt{d_k}}\right)\mathbf{V}_h, \\
\mathbf{Q}_h = \mathbf{E} \mathbf{W}_h^Q, \quad \mathbf{K}_h = \mathbf{Z}_v \mathbf{W}_h^K, \quad \mathbf{V}_h = \mathbf{Z}_v \mathbf{W}_h^V,
\end{gathered}
\end{equation}
where $\mathbf{W}_h^Q$, $\mathbf{W}_h^K$, $\mathbf{W}_h^V$, and $\mathbf{W}^O$ are learnable parameters.

To summarize the above process, the regional image features $\mathbf{Z}_v$ output by the visual encoder and the disease entity name embeddings $\mathbf{E}$ are cross-attended and passed through a three-layer GCN whose adjacency matrix is defined by our knowledge graph to output the disease-related feature $\mathbf{Z}_g=\mathbf{N}^3$.

\subsection{Feature Fusion}
After obtaining the regional features $\mathbf{Z}_v \in \mathbb{R}^{S \times d_v}$, and the knowledge-enhanced disease-related features $\mathbf{Z}_g \in \mathbb{R}^{M \times d_v}$, we employed a multi-head attention network to align their dimensions by using $\mathbf{Z}_v$, $\mathbf{Z}_g$ and $\mathbf{Z}_g$ as the query ($\mathbf{Q}$), key ($\mathbf{K}$) and value ($\mathbf{V}$). The attention output, denoted as $\tilde{\mathbf{Z}}_g \in \mathbb{R}^{S \times d_v}$, shares the same dimensions as $\mathbf{Z}_v$. In the following, we propose two fusion strategies designed to integrate these two types of features. \\

\noindent\textbf{Element-wise Fusion}~~~This approach uses an element-wise weighted sum for the final integrated feature representation, employing a trainable gate to determine the importance of each element in the two feature types.
The fused features $\mathbf{Z}_f \in \mathbb{R}^{S \times d_v}$ is obtained by:
\begin{equation} \label{equ: element fusion}
\begin{gathered}
\mathbf{Z}_f = \textbf{gate} \odot \mathbf{Z}_v + (1-\textbf{gate}) \odot \tilde{\mathbf{Z}}_g, \\
\textbf{gate} = sigmoid([\mathbf{Z}_v; \tilde{\mathbf{Z}}_g] \cdot \mathbf{W}^g),
\end{gathered}
\end{equation}
\noindent where $[\mathbf{Z}_v; \tilde{\mathbf{Z}}_g]$ represents the concatenation of $\mathbf{Z}_v$ and $\tilde{\mathbf{Z}}_g$, and $\mathbf{W}^g$ is a learnable parameter. The operation $\odot$ signifies element-wise multiplication. \\

\noindent\textbf{Modality-wise Fusion}~~~Inspired by the Mixture of experts (ME)~\cite{masoudnia2014mixture}, we designed two distinct expert networks: one to process disease-related features $\tilde{\mathbf{Z}}_g$ and the other for general regional features $\mathbf{Z}_v$. To dynamically allocate the contribution of each expert's output, we put forward a soft router module, represented by $\mathbf{G}(x)$, functioning as a gating network. This gate is implemented as a multi-layer perceptron (MLP). Unlike the element-wise fusion operating at the individual element level, modality-wise fusion treats each feature set as an integral unit for combination. 
The combined output is formulated as:
\begin{equation} \label{equ: modality fusion}
\begin{gathered}
\mathbf{Z}_f = g_1 E_1(\mathbf{Z}_v) + g_2 E_2(\tilde{\mathbf{Z}}_g), ~~~
[g_1, g_2] = \mathbf{G}(\mathbf{Z}_v, \tilde{\mathbf{Z}}_g).
\end{gathered}
\end{equation}
Here, $E_1$ and $E_2$ denote the expert networks comprising of a linear layer and layer normalization. The soft router $\mathbf{G(x)}$ assesses and decides the relevance of each expert ($E_1$ and $E_2$) 
in the fusion process. The weights $g_1$ and $g_2$ are computed as probability values, indicating the importance assigned to each expert's output in the final feature representation $\mathbf{Z}_f$.

\subsection{Report generation}
We employ LLaMA2-7B to generate radiology reports, leveraging the fused features as the visual prompt. Our instruction prompt is designed following the template of LLaMA2. Given a set of fused features $\mathbf{Z}_f$ output by the feature fusion module according to Eqn.~\ref{equ: modality fusion}, our prompt is designed as: \textit{`[INST] $\mathbf{Z}_t$ <feats> $\mathbf{Z}_f$ </feats> [/INST].'}, where $\mathbf{Z}_t$ is a constant instruction text: \textit{`Generate a comprehensive and detailed diagnosis report for this radiology image.'}. Before input to LLaMA2, all text words in the prompt are tokenized and embedded by LLaMA's tokenizer and word embedding layers. Recall that $\mathbf{X}_v$ denotes the input image. Our overall model is optimized by minimizing the cross-entropy loss:
\begin{equation}
\mathcal{L}_{CE}(\theta) = -\sum_{i=1}^{N_r} \log p_\theta (t_i^{*}|\mathbf{X}_v,\mathbf{Z}_t,t_{1:i-1}^{*}),
\end{equation}
where $\theta$ denotes model parameters, and $t_i^{*}$ is the $i$-th word in the ground truth report with a length of $N_r$ words.

\section{Experiments}
\noindent\textbf{Datasets}~~~Two widely used benchmarks are involved in our experiments.

\noindent\underline{IU-Xray} is from Indiana University Chest X-ray Collection (IU-Xray)~\cite{demner2016preparing}, comprising 3,955 radiology reports linked to 7,470 chest X-ray images. Following the partitioning guidelines of~\cite{chen2020generating}, we divided the dataset into training, testing, and validation sets with a ratio of 7:1:2. 

\noindent\underline{MIMIC-CXR}~\cite{johnson2019mimic} comprises 377,110 chest X-ray images and 227,835 associated reports from 64,588 patients at the Beth Israel Deaconess Medical Center (2011-2016). 
For consistency and fair comparison, we utilized the dataset's division defined by~\cite{chen2020generating}, i.e., 270790 images for training and 3858 for testing.\\

\noindent\textbf{Implementation Details}~~~In this work, we employed LLaMA2-7B~\footnote{https://huggingface.co/meta-llama/Llama-2-7b-chat-hf} as the LLM and Swin Transformer~\footnote{https://huggingface.co/microsoft/swin-base-patch4-window7-224} as the visual encoder. We used 3 layers GCN to aggregate the disease-related features through the medical domain knowledge graph. The model was trained on two NVIDIA A6000 48GB GPUs, utilizing a mini-batch size of 8 and a learning rate of 1e-4. For testing, a beam search strategy was adopted with a beam width of 3 to balance between computational efficiency and output quality.\\


\noindent\textbf{Evaluation Metrics}~~~We used traditional natural language generation (NLG) metrics (e.g., BLEU~\cite{papineni2002bleu}, ROUGE-L~\cite{lin2004rouge}, METEOR~\cite{banerjee2005meteor}, and CIDEr~\cite{vedantam2015cider}), as well as recent clinic-related metrics RadGraph F1~\cite{jain2021radgraph} and BERTScore~\cite{zhang2019bertscore}, following insights from~\cite{yu2023evaluating}. The latter offers a closer alignment with radiologist assessments than NLG metrics and the Chexpert~\cite{irvin2019chexpert} clinical efficacy score~\cite{yu2023evaluating}. Moreover, we incorporated the RadCliQ metric~\cite{yu2023evaluating}, a comprehensive measure that combines individual metrics to better correlate with radiologist evaluations~\footnote{https://github.com/rajpurkarlab/CXR-Report-Metric/tree/v1.1.0}.\\

\begin{table}[t]
\centering
\caption{Comparison on MIMIC-CXR and IU-Xray datasets. The highest scores are highlighted in bold, the second-highest scores are indicated with an underline.}
\label{Table:ComparisonWithSOTA}
\resizebox{\linewidth}{!}{
\begin{tblr}{
  cell{0-28}{3-9} = {c},
  vline{2,3} = {-}{},
  hline{1-2,15,27} = {-}{},
  hline{14,26} = {2-9}{},
}
Dataset             & Methods                                         & BLEU-1         & BLEU-2         & BLEU-3         & BLEU-4         & ROUGE          & METEOR         & CIDEr \\
                    & Show-Tell~\cite{xu2015show}                     & 0.308          & 0.190          & 0.125          & 0.088          & 0.256          & 0.122          & 0.096 \\
                    & AdaAtt~\cite{lu2017knowing}                     & 0.314          & 0.198          & 0.132          & 0.094          & 0.267          & 0.128          & 0.131 \\
                    & M2Transformer~\cite{cornia2020meshed}           & 0.332          & 0.210          & 0.142          & 0.101          & 0.264          & 0.134          & 0.142 \\
                    & R2Gen$^\dagger$~\cite{chen2020generating}       & 0.353          & 0.218          & 0.145          & 0.103          & 0.277          & 0.142          & -     \\
                    & R2GenCMN$^\dagger$~\cite{chen2022cross}         & 0.353          & 0.218          & 0.148          & 0.106          & 0.278          & 0.142          & -     \\
MIMIC-CXR           & PPKED$^\dagger$~\cite{liu2021exploring}         & 0.36           & 0.224          & 0.149          & 0.106          & 0.284          & 0.149          & 0.237 \\
                    & GSK$^\dagger$~\cite{yang2022knowledge}          & 0.363          & 0.228          & 0.156          & 0.115          & 0.284          & -              & 0.203 \\
                    & MSAT$^\dagger$~\cite{wang2022medical}           & 0.373          & 0.235          & 0.162          & 0.120          & 0.282          & 0.143          & 0.299 \\
                    & METransformer$^\dagger$~\cite{wang2023metransformer}      & 0.386          & 0.250          & 0.169          & 0.124          & 0.291          & 0.152          & \underline{0.362} \\
                    & CvT2DistilGPT2$^\dagger$~\cite{nicolson2023improving}     & 0.393          & 0.248          & 0.171          & 0.127          & -              & 0.155          & \textbf{0.389} \\
                    & RaDialog-RG$^\dagger$~\cite{pellegrini2023radialog}       & 0.346          & -              & -              & 0.095          & 0.271          & 0.140          & - \\
                    & R2GenGPT$^\dagger$~\cite{wang2023r2gengpt}            & \underline{0.411}          & \underline{0.267}          & \underline{0.186}          & \underline{0.134}          & \underline{0.297}        & \underline{0.160}         &  0.269 \\ 
                    & Ours (Modality-wise Fusion)                           & \textbf{0.417}             & \textbf{0.274}             & \textbf{0.192}             & \textbf{0.140}             & \textbf{0.305}            & \textbf{0.165}           &  0.289 \\             
                    & Show-Tell~\cite{xu2015show}                   & 0.243          & 0.130          & 0.108          & 0.078          & 0.307          & 0.157          & 0.197          \\
                    & AdaAtt~\cite{lu2017knowing}                   & 0.284          & 0.207          & 0.150          & 0.126          & 0.311          & 0.165          & 0.268          \\
                    & M2transformer~\cite{cornia2020meshed}         & 0.402          & 0.284          & 0.168          & 0.143          & 0.328          & 0.170          & 0.332          \\
                    & R2Gen$^\dagger$~\cite{chen2020generating}     & 0.470          & 0.304          & 0.219          & 0.165          & 0.371          & 0.187          & -              \\
                    & R2GenCMN$^\dagger$~\cite{chen2022cross}       & 0.475          & 0.309          & 0.222          & 0.170          & 0.375          & 0.191          & -              \\
IU-Xray             & KERP$^\dagger$~\cite{li2019knowledge}         & 0.482          & 0.325          & 0.226          & 0.162          & 0.339          & -              & 0.280           \\
                    & PPKED$^\dagger$~\cite{liu2021exploring}       & 0.483          & 0.315          & 0.224          & 0.168          & 0.376          & 0.190          & 0.351           \\
                    & MSAT$^\dagger$~\cite{wang2022medical}         & 0.481          & 0.316          & 0.226          & 0.171          & 0.372          & 0.190          & 0.394          \\
                    & METransformer$^\dagger$~\cite{wang2023metransformer}       & 0.483          & \underline{0.322}          & \underline{0.228}          & 0.172          & \underline{0.380}          & 0.192          & 0.435          \\
                    & CvT2DistilGPT2$^\dagger$~\cite{nicolson2023improving}      & 0.473                   & 0.304              & 0.224                   & \underline{0.175}  & 0.376             & 0.200                   & \textbf{0.694} \\
                    & R2GenGPT$^\dagger$~\cite{wang2023r2gengpt}                 & \underline{0.488}       & 0.316              & \underline{0.228}       & 0.173              & 0.377             & \underline{0.211}       & 0.438 \\
                    & Ours (Modality-wise Fusion)                                & \textbf{0.490}          & \textbf{0.323}     & \textbf{0.232}          & \textbf{0.180}     & \textbf{0.385}    & \textbf{0.218}          & \underline{0.491} \\
\end{tblr}
}
\end{table}  

\noindent\textbf{Comparison with the state-of-the-art}~~~Table~\ref{Table:ComparisonWithSOTA} compares KARGEN's performance with state-of-the-art (SOTA) methods in image captioning and report generation on the MIMIC-CXR and IU-Xray datasets. Table~\ref{Table:ComparisonWithSOTA_Rad} focuses on comparisons using the metrics RadGraph F1, Bert score, and RadCliQ. Except those $\dagger$ marked methods whose performances are quoted from their respective papers, we re-run publicly released codes of comparison methods on the same training-test partition as our approach.

\begin{table}[h]
\centering
\caption{Evaluation of Clinic-related Metrics on MIMIC-CXR}
\label{Table:ComparisonWithSOTA_Rad}
\resizebox{\linewidth}{!}{
\begin{tblr}{
  cell{0-8}{2-4} = {c},
  vline{2} = {-}{},
  hline{1-2,7-8} = {-}{},
}
Methods                                                     & RadGraph F1($\uparrow$)         & Bert Score($\uparrow$)         & RadCliQ($\downarrow$)  \\
R2Gen~\cite{chen2020generating}                             & 0.172                           & 0.406                          & 1.228 \\          
R2GenCMN~\cite{chen2022cross}                               & 0.182                           & 0.418                          & 1.182 \\
CvT2DistilGPT2~\cite{nicolson2023improving}                 & 0.196                           & 0.374                          & 1.220 \\
RaDialog-RG$^\dagger$~\cite{pellegrini2023radialog}         & -                               & 0.40                           & -     \\
R2GenGPT~\cite{wang2023r2gengpt}                            & 0.187                           & 0.415                          & 1.207 \\ 
Ours (Modality-wise Fusion)                                 & \textbf{0.203}                  & \textbf{0.421}                 & \textbf{1.165}
\end{tblr}
}
\end{table}  

As seen in Table~\ref{Table:ComparisonWithSOTA}, KARGEN outperforms existing methods across almost all evaluation metrics on both datasets. Specifically, it surpasses both traditional image captioning methods such as Show-Tell~\cite{xu2015show} and M2Transformer~\cite{cornia2020meshed}, advanced transformer-based R2Gen methods such as METransformer~\cite{wang2023metransformer} and PPKED~\cite{liu2021exploring}, and very recent LLM-based models like CvT2DistilGPT2~\cite{nicolson2023improving}, RaDialog-RG~\cite{pellegrini2023radialog}, and R2GenGPT~\cite{wang2023r2gengpt} in nearly all metrics. On MIMIC-CXR, our BLEU-4 score sees a noteworthy improvement of 4.5\%, rising from 0.134 to 0.140. Although our CIDEr score of 0.289 is lower than that of METransformer (0.362) and CvT2 (0.389), this discrepancy can be attributed to the employment of a unique expert voting in METransformer and the utilization of a larger image size (384x384 pixels) in CvT2. On IU-Xray, KARGEN consistently shows promising performance. In addition to NLG metrics, it is more important to see KARGEN achieve the highest scores in the clinic-related metrics RadGraph F1, Bert Score, and RadCliQ, reinforcing its advantages. This significant advancement is attributed to the integration of disease-related features, enhancing the model's ability to accurately identify diseases. It is noted that RaDialog-RG~\cite{pellegrini2023radialog} constructed prompts using the output of a trained disease classifier to incorporate disease information. Compared with it, our disease knowledge graph could carry more complicated disease relationships to assist LLMs for R2Gen.\\

\begin{figure}[t]
\centering
\includegraphics[width=\textwidth]{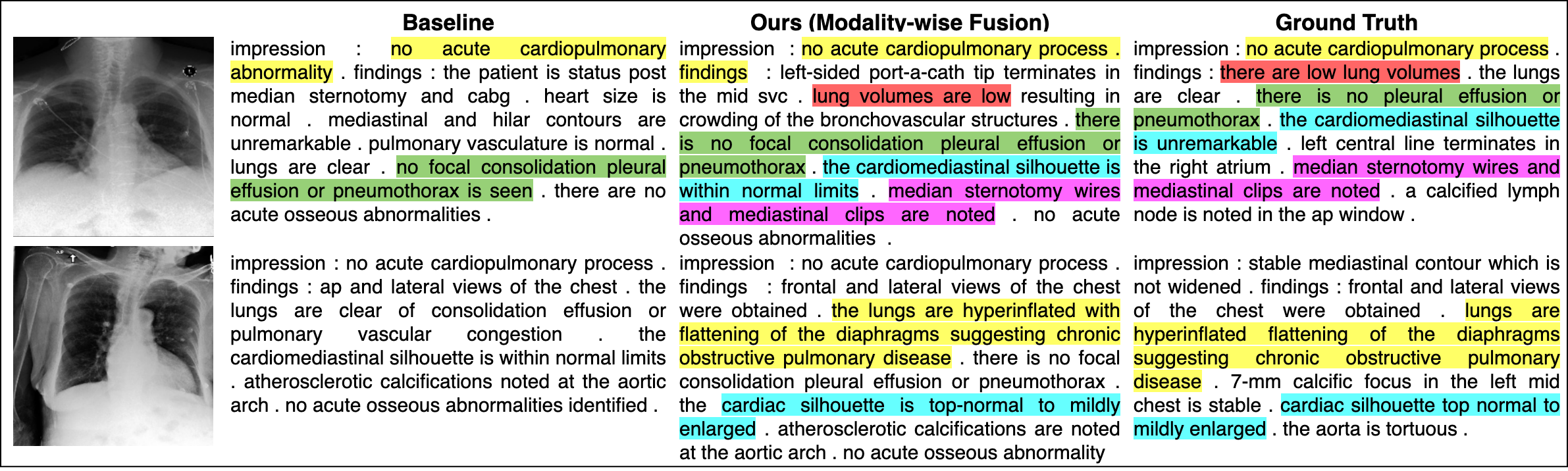}
\caption{Examples of the generated reports. For better illustration, different colours highlight different medical terms in the reports. 
} 
\label{Figs: case}
\end{figure}

\noindent\textbf{Ablation Study:} Table~\ref{tab:ablation} summarizes our ablation study on the MIMIC-CXR dataset, singling out the contribution of each component, including knowledge-enhanced disease-related features, Graph Convolutional Network (GCN), and fusion methods. 
As seen, utilizing only regional or disease-related features yields moderate performance, while integrating both significantly enhances model effectiveness. Modality-wise fusion appears to be a superior fusion strategy. 
Examining configurations excluding GCN, which aggregates features through the graph, indicates less pronounced performance gains. Our complete model yields notably more accurate and descriptive outcomes compared to the baseline. Fig.~\ref{Figs: case} shows examples of generated reports. As seen, our model effectively captures both normal and abnormal contents consistent with the ground truth, while the baseline fails to generate the contents marked in red and magenta colors, confirming the benefits of our integration of knowledge-enhanced disease-related features via modality-wise fusion.
\begin{table}[h]
\centering
\caption{Ablation study. $\mathbf{Z}_v$ is for regional features, and $\tilde{\mathbf{Z}}_g$ for disease-related features. \textbf{E}, \textbf{M}, and \textbf{A} stand for Element-wise, Modality-wise and Average.}
\label{tab:ablation}
\resizebox{\linewidth}{!}{
\begin{tblr}{
  cell{0-27}{2-10} = {c},
  vline{2,8} = {-}{},
  hline{1-2,8} = {-}{},
}
Dataset     & $\mathbf{Z}_v$      & $\tilde{\mathbf{Z}}_g$       & \textbf{E}-Fusion                & \textbf{M}-Fusion        & \textbf{A}-Fusion    & GCN            & BLEU-4         & ROUGE          & METEOR         & CIDEr          \\
            & \checkmark          &                              &                         &                 &             &             & 0.134          & 0.297          & 0.160          & 0.269          \\
            &                     & \checkmark                   &                         &                 &             & \checkmark  & 0.134          & 0.302          & 0.160          & 0.259          \\
            & \checkmark          & \checkmark                   &                         &                 & \checkmark  & \checkmark  & 0.132          & 0.303          & 0.156          & 0.245           \\
MIMIC-CXR   & \checkmark          & \checkmark                   & \checkmark              &                 &             & \checkmark  & 0.137          & 0.303          & 0.163          & 0.281          \\
            & \checkmark          & \checkmark                   &                         & \checkmark      &             &             & 0.134          & 0.301          & 0.162          & 0.270          \\
            & \checkmark          & \checkmark                   &                         & \checkmark      &             & \checkmark  & \textbf{0.140}          & \textbf{0.305}          & \textbf{0.165}          & \textbf{0.289}          \\
\end{tblr}
}
\end{table}

\section{Conclusions}
In this paper,  we propose a novel framework integrating LLMs with a medical knowledge graph for R2Gen. Our work highlights the value of incorporating disease-specific knowledge graphs with LLMs and the importance of fusing regional image features with knowledge-enhanced disease-related features to improve the quality and clinic utility of the generated reports. In the future, larger knowledge graphs will be explored along this line.

\begin{credits}

\subsubsection{\discintname}
The authors have no competing interests to declare that are relevant to the content of this article.
\end{credits}
%
%
%

\bibliographystyle{splncs04}
\bibliography{paper-0877}
%




\end{document}